\documentclass[11pt,a4paper]{article}
\usepackage[pdftex]{graphicx}
\usepackage[pdftex]{color}
\usepackage{subfigure}
\usepackage[square]{natbib}
\bibpunct[:]{(}{)}{;}{a}{}{:}
\usepackage{amssymb}
\usepackage{color}

\newcommand{\eqref}[1]{Formula~ \ref{#1}}
\newcommand{\secref}[1]{\S \ref{#1}}
\newcommand{\figref}[1]{Figure~\ref{#1}}
\newcommand{\tabref}[1]{Table~\ref{#1}}
\newcommand{\figheight}{4.5cm}

\newcommand{\allfig}[1]{
\centering
\tiny
\subfigure[Zipf's Law]
{\includegraphics[height=\figheight]{figs/#1_Zipf.png}}
\hspace*{0.3cm}
\subfigure[Heaps' Law]
{\includegraphics[height=\figheight]{figs/#1_Heap.png}}
\hspace*{0.3cm}
\subfigure[Ebeling's Method]
{\includegraphics[height=\figheight]{figs/#1_chars_scaling.png}}
\hspace*{-0.3cm}
\subfigure[Taylor's Law]
{\includegraphics[height=\figheight]{figs/#1_taylor.png}}
\hspace*{-0.2cm}
\subfigure[Long-Range Corr.]
{\includegraphics[height=\figheight]{figs/#1_acf.png}}
}

\usepackage{url}

\title{Assessing Language Models with Scaling Properties}
\author{Shuntaro Takahashi\textsuperscript{1\dag},
Kumiko Tanaka-Ishii\textsuperscript{2\ddag *}
\\
\textsuperscript{1\dag} The University of Tokyo, Graduate School of Frontier Sciences
\\
\textsuperscript{2\ddag *} The University of Tokyo, Research Center for Advanced Science and Technology
}
\begin{document}
\maketitle

\begin{abstract}
Language models have primarily been evaluated with perplexity. While perplexity 
quantifies the most comprehensible prediction performance, it does not provide 
qualitative information on the success or failure of models. Another approach for 
evaluating language models is thus proposed, using the scaling properties of natural 
language. Five such tests are considered, with the first two accounting for the 
vocabulary population and the other three for the long memory of natural language. The 
following models were evaluated with these tests: $n$-grams, probabilistic context-free 
grammar (PCFG), Simon and Pitman-Yor (PY) processes, hierarchical PY, and neural 
language models. Only the neural language models exhibit the long memory properties 
of natural language, but to a limited degree. The effectiveness of every test of these 
models is also discussed.
\end{abstract}

\section{Introduction}
The performance of language models has generally been evaluated with perplexity 
\citep{perplexity}, which quantifies the predictive power of such models. Since the 
development of $n$-gram models, state-of-the-art neural language models have radically reduced 
perplexity. Although perplexity is easy to compute and its signification is 
comprehensible, we consider it to have two main drawbacks.
First, perplexity does not provide information on how a model is limited with 
respect to the important linguistic aspects, such as generalization of word concepts, 
syntactic structure, and long-term dependency. Although perplexity indicates the overall 
performance across various aspects, we want to know more specifically which aspects 
are challenging for the language models. 
Second, we cannot obtain a lower bound of perplexity in a dataset and 
thus cannot quantitatively recognize the difference of a model from natural language or an {\em ideal} 
language model that perfectly reproduces natural language.

To compensate for these drawbacks of perplexity, we propose to assess language models 
with a set of scaling properties exhibited by natural language. The assessment is 
conducted by investigating whether the text generated by a language model would 
exhibit these scaling properties.  We presented this scheme in \citep{Takahashi_2017} in which the reproducibility of the character level neural language models are investigated.  This assessment has two advantages over the perplexity measure.
First, since the scaling properties are designed to measure some
aspect underlying a sequence, they can provide information on how
limited a model is in terms of that aspect. Second, since the scaling
properties quantify the behavior of a data set in terms of exponents,
we can evaluate how much the text generated from a model differs from
natural language.

The scaling properties in this article are roughly categorized into two types: those 
related to word frequency distribution and those related to long memory. For the first 
category, we consider the well-known Zipf's law and Heaps' law. For the second 
category, we consider properties that quantify the memory underlying a sequence. We 
then test whether texts generated from various language models exhibit every property.

\begin{figure*}[t]
\allfig{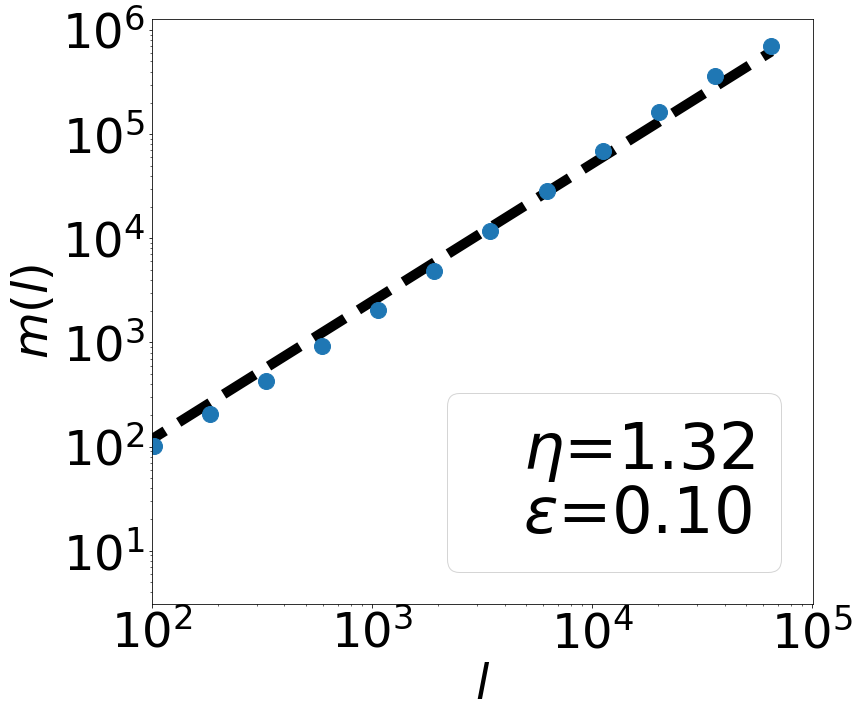}
\vspace*{-0.5cm}
\caption{Five scaling properties for wikitext-2 (WT2). \label{fig:phy}}
\end{figure*}  

\section{Related Work} 

Many studies of language models have considered Zipf's and Heaps'
laws. The laws concern the word frequency distribution,  
characterizing how a large part of vocabulary
consists of rare words, causing problems of sparseness and unknown
words.  The Chinese restaurant process, as a kind of PY process
\citep{pitman}, was introduced partly because it satisfies Zipf's and
Heaps' law \citep{goldwater_ml,teh06}. 
In the recent works on neural language models, 
application of the two laws was considered to improve the
architecture. \citet{Jozefowicz_2016} evaluated 
models in terms of word frequencies and concluded that a neural
model is more capable of predicting rare words than is a
Kneser-Ney language model \citep{Kneser_1995}. 
\citet{Merity_2016} constructed a new dataset, {\em wikitext-2}
(WT2) for processing rare words with respect to the word frequency distribution, and it has become a
standard dataset.

Long-range dependency has mainly been explored in 
studying the architecture of neural networks. After the first neural language model was 
proposed \citep{Bengio_2003}, \citet{Mikolov_2010} first introduced recurrent neural 
networks (RNNs) to language modeling, with the potential to maintain long-term 
dependency. In more recent works, \citet{Merity_2016} introduced pointer 
networks \citep{Vinyals_2015} to the task and demonstrated that they 
improve perplexity on the regular long short-term memory (LSTM) language model.
\citet{Grave_2016} integrated a cache model designed to sample from context to model the 
observation of the clustering behavior of certain words. 
\citet{Merity_2017} investigated the effect of a cache model in terms of the difference 
in log perplexity of each word. These analysis methodologies for long-term dependency, 
however, have not been formalized and instead rely on case studies. 
\citet{Lin_Tegmark_2016} investigated how an LSTM-like architecture 
could reproduce the power decay of mutual information. 
Although their argument is theoretically valid, natural language 
does not exhibit power decay of mutual information. Our interest is therefore to 
introduce better measures of the long-range dependency of natural language.

\section{Scaling Properties}
\label{scaling}
This section provides a summary of the scaling properties of natural language and 
defines how to evaluate word or character sequences with them. Nine scaling laws of 
natural language are acknowledged \citep{altmann16}. Some of them deal with word 
formation and network structure and do not directly relate to language modeling. This 
leaves five other scaling properties that are still mathematically closely related. We can 
roughly categorize them into those based on vocabulary population and long memory. 
The following subsections proceed through the example of applying these scaling 
properties with WT2 \citep{Merity_2016}, as shown in \figref{fig:phy}.

We test language models by generating text from them and evaluating the scaling 
properties. In this article, there are two levels of judgment for these properties.
\begin{description}
\item[Q1] Does the scaling property hold qualitatively?
\item[Q2] How different is the exponent?
\end{description}
As revealed in the following subsections, many language models fail to satisfy even the 
first criterion for some properties. For models that satisfy Q1, their exponents are 
estimated and compared with those of the original text.
Consider a power law $y \propto z^{\kappa}$ for points
($y_1$,$z_1$),$\ldots$,($y_N$,$z_N$). The exponent $\kappa$ is
estimated by the least-squares method, i.e., $\hat{\kappa} = \arg
\min_{\kappa} \varepsilon(\kappa)$, where $\varepsilon(\kappa)\equiv
\sqrt{\sum_{i=1}^N (\log y_i - \log z_i^{\kappa})^2/N}$. The error
reported here is the average error per point, i.e.,
$\varepsilon(\hat{\kappa})$.

\subsection{Zipf's Law and Heaps' Law}
\label{sec:zipf}

Given a text, let $r$ be the rank of a particular word type and $f(r)$ be its frequency. 
The well-known Zipf's law formulates a power-law relation between the rank and 
frequency:
\begin{equation}
f(r) \propto r^{-\alpha}, \hspace*{1cm} \alpha \approx 1.0. \label{eq:zipf}
\end{equation}
In fact, this scaling generally holds not only for unigrams but also for
$n$-grams, with smaller $\alpha$. The first left graph in
\figref{fig:phy} shows Zipf distributions for WT2, with unigrams in
red and bigrams in blue. Because WT2 replaces rare words having
frequencies under a certain threshold with \verb!<unk>!, the tail of the
unigram distribution disappears. The Zipf distributions for unigrams
and bigrams typically cross in the middle.
In reality, plots for real natural language texts are often not
aligned linearly, making the exponent difficult to estimate. Previous
works have dealt with this problem, but it is beyond the scope of this
article. In this work, we therefore do not estimate $\alpha$ either.

Heaps' law is another scaling property and shows how vocabulary grows with text size, 
forming a power-law distribution. Let $n$ be the length of a text and $v(n)$ be its 
vocabulary size. Then Heaps' law is formulated as the following relation:
\begin{equation}
 v(n) \propto n^{\beta}, \ \  0 < \beta < 1. \label{eq:heaps}
\end{equation}
The second graph in \figref{fig:phy} shows the result for WT2. The exponent here is 
0.75, which is smaller than 1.0 (dashed black line), with $\varepsilon=0.13$.
Because of the replacement of unknown words in WT2, the plot exhibits convex growth. 
Also, note that Heaps' law can be deduced from Zipf's law 
\citep{BaezaYates_2000,Leijenhorst_2005,Lu_2010}.

\subsection{Long Memory}
The statistical mechanics domain has introduced two directions for considering long 
memory: fluctuation analysis and long-range correlation. Here, we 
introduce two fluctuation analysis methods, one for characters and one for words, and 
one long-range correlation method, applied to words. Although these methods are 
related analytically for a well-formed time series, for real phenomena with 
finite-size samples, the relations are non-trivial. The generated texts could behave 
similar to a random sequence or contrarily exhibit better long memory than a natural 
language text.

\subsubsection{Fluctuation Analysis}
Fluctuation analysis quantifies the strength of memory and the degree of symbol 
occurrence burstiness underlying a text.

\subsubsection*{Ebeling's Method}

Burstiness has been known to occur in various natural and social
domains. Fluctuation analysis originated in \citep{hurst}, motivated by
the need to quantify the degree of burstiness of Nile River
floods. The method applies only for numerical series, so
\citep{montemurro} applied it for texts by
transforming a word sequence into a rank sequence, which obscures the
results.

One work \citep{Ebeling1994} applied a simple fluctuation analysis method on text. 
That work showed how the variance of characters grows by a power law with respect to 
the text length (i.e., time span). Given a set of elements $|W|$ (characters in this method), 
let $y(k,l)$ be the number of the $c_k \in W$ within text length $l$. Then,
\begin{equation}
  m(l) = \sum_{k=1}^{|W|} m_2(k,l) \propto l^\eta,  
\end{equation}
where $m_2(k,l)$ is the variance of $y(k,l)$:
\begin{equation}
  m_2(k,l) = <y^2(k,l) > - (< y(k,l)>)^2.
\end{equation}
Theoretically, if the time series is independent and identically distributed (i.i.d.), then 
$\eta=1.0$. \citep{Ebeling1994} showed that the Bible has $\eta = 1.69$, thus 
exhibiting larger fluctuation than an i.i.d. sequence. This article successfully 
demonstrates that a character sequence has long memory. The third graph in 
\figref{fig:phy} shows $m(l)$ for WT2. The exponent is 1.32 with $\varepsilon=0.10$. 

\subsubsection*{Taylor's Law}
While Ebeling's method considers the growth of variance with respect to a subsequence 
of length $l$, Taylor's method considers that growth with respect to the mean within 
length $l$. Because the mean of a subsequence linearly correlates with $l$, the two 
methods are closely related. While Ebeling's method sums the variance over all 
elements of $W$, Taylor's method estimates the scaling exponent from the points of all words. 
Because of this difference, Ebeling's method is not applicable for detecting long 
memory underlying a word sequence: the exponent becomes 1.0 in this case, like an 
i.i.d. sequence. In general, Taylor's law is more robust and is applicable to words and 
also to characters, if for a large set size.

Taylor's law was originally reported in two pioneering works 
\citep{taylor-nature,taylor-smith} and has been applied in various domains as reported in 
\citep{taylor}. Its application for language data, however, has been scarce. The only such 
study so far is \citep{altmann14-taylor}. However the work considered vocabulary size rather 
than word occurrence, and its differences from the original Taylor analysis make theoretical interpretation inapplicable. 
We introduce another method proposed by \citet{Kobayashi_2018} with which we can quantify the long range dependence of natural language and symbolic time-series in general and is highly interpretable.

Given a text produced from a set of words, $W$, for a given segment size $l$, the 
number of occurrences of a particular word $w_k \in W$ is counted and the mean 
$\mu_k$ and standard deviation $\sigma_k$ are calculated for $w_k$. Doing this for 
all elements of $W$ gives the distribution of $\sigma$ with respect to $\mu$. Taylor's 
law then holds when $\sigma$ and $\mu$ are correlated by a power law:
\begin{equation}
  \sigma  \propto \mu^\zeta \\. 
\end{equation}
Experimentally, the Taylor exponent $\zeta$ is known to range over $0.5 \leq \zeta 
\leq 1.0$.

The advantage of Taylor's law over the other analysis methods for long memory is the 
interpretability of the exponent $\zeta$. The two limit values $\zeta=0.5,1.0$ provide a 
clear interpretation. For an i.i.d. process, it is proved that $\zeta=0.5$. For a sequence in 
which all segments of length $l$ contain the same proportions of the elements of $W$, 
$\zeta=1.0$. For example, given $W=\{a, b\}$, suppose that $b$ always occurs twice 
as often as $a$ in all segments (e.g., one segment with three $a$ and six $b$, another 
segment with one $a$ and two $b$, etc.). Then, both the mean and standard deviation 
for $b$ are twice those for $a$, and therefore $\zeta=1.0$. The Taylor exponent thus 
quantifies how consistently words co-occur in texts. For this reason, a set of tweets from 
the same source or CHILDES texts would typically exhibit a higher exponent $\zeta$.

The fourth graph in \figref{fig:phy} shows the result for WT2 with 
$l=5620$ ($l$ can be any value larger than one). The plot exhibits a power law, 
although some deviation from the regression line is also visible. The Taylor exponent is 
$\zeta=0.62$, with $\varepsilon=0.15$.

\subsubsection{Long-Range Correlation}
Burstiness of word occurrence leads to a related observation: how subsequences are 
similar. Such thinking resulted in another genre of analysis methods, called {\em 
long-range correlation analysis}, which measures the self-similarity within two 
subsequences of a time series.

A time series is said to be long-range correlated if the correlation function $c(s)$ for 
two subsequences separated by distance $s$ follows a power law:
\begin{equation}
 c(s) \propto s^{-\xi}, \ \ s > 0, 0 < {\xi}<1.
\end{equation}
A widely used choice for $c(s)$ is the autocorrelation function (ACF). By using the 
ACF, the value of $c(s)$ ranges between -1 and 1. When a sequence is long-range 
correlated, $c(s)$ takes positive values for $s$ until about 1/100 of the length 
\citep{Lennartz2009}, whereas $c(s)$ fluctuates around zero for a sequence without 
temporal correlation.

Since the ACF is applicable only for numerical time series, application of this method 
for natural language requires transforming a natural language text into a numerical time 
series. Among recent reports \citep{Altmann2009,Altmann2012,plos16}, we apply the 
most recent, a rare word clustering method with parameter $Q=16$ \citep{plos16}.

Application of long-range correlation analysis to word sequences in
WT2 produced the last graph in \figref{fig:phy}. As noted in the
legend, $\xi =0.33$ and $\varepsilon=0.04$
with $c(s)$ is all positive up to
1/100 of the sequence length.
Throughout this article, for $\varepsilon$ of this metric only, 
it is measured for $s \leq 100$.

\section{Language Models}
\label{sec:models}
This section explains the language models used in this article. Each model defines a 
probability distribution $P(x_{t+1})$, given $X_{1}^{t} = x_1,x_2,\ldots,x_t$. We 
probabilistically generated texts following the output distributions of these models.
The all graphs for the scaling properties of the language models are available as supplementary material.

\subsection{$N$-gram Model}
An $n$-gram language model is the most basic model, as it is a Markov model of order 
$n-1$.
This article considers a 3-gram model and 5-gram model with back-off.  
Note that the perplexity of
the models here should tend to be lower than in previous reports, as
we did not adopt the \verb!<BOS>! and \verb!<EOS>! tags to maintain inter-sentence structure.

\subsection{Grammatical Model}
Probabilistic context-free grammar (PCFG) is the most basic grammatical model. We 
constructed this grammar model with the annotated {\it Penn Tree Bank} (PTB) dataset and 
used the Natural Language Toolkit (NLTK) to generate sentences according to the 
probabilities assigned to productions. Unlike an $n$-gram model, a PCFG ensures the 
grammatical correctness of all productions.

\begin{figure*}[t]
  \allfig{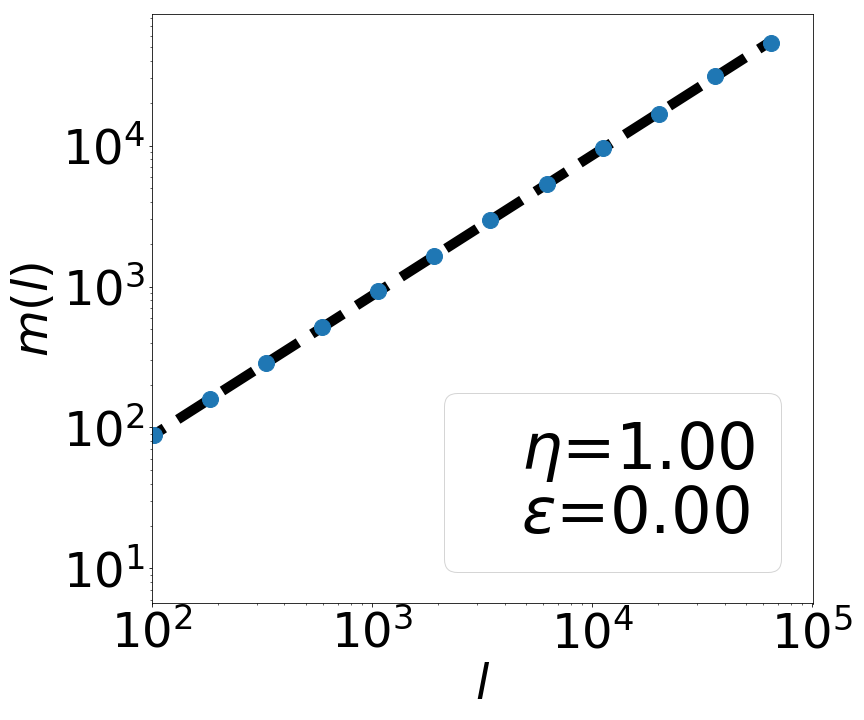}
  \vspace*{-0.5cm}
\caption{Scaling properties of HPYLM trained with WT2. \label{fig:hpylm}}
\end{figure*}

\subsection{Simon/Pitman-Yor Models}
\label{sec:py}
The Simon and Pitman-Yor (PY) processes are important for our perspective, because 
they are capable of reproducing Zipf's law and Heaps' law with simple formulations. It 
is thus interesting to see whether they satisfy the other scaling properties.

These are generative models, and a sequence is formulated over time, either through (1) 
introduction of new words or (2) sampling from the past sequence. Let 
$K(X_{1}^{t})$ be the number of word types existing in $X_{1}^{t}$, and let 
$n_k(X_{1}^{t})$ be the frequency of the $k$th word type in $X_1^t$. The sequence 
starts with $K(X_{0})=1$ and $X_{0} = x_0$ at $t = 0$.

For $t \geq 1$, given a constant $a$ with $0 < a < 1$, the Simon process 
\citep{simon55} introduces a new word with probability $a$, or a word is sampled from 
$X_1^t$ with probability $1-a$:
$$
\footnotesize
P(x_{t+1}=w_k) = \left\{
\begin{array}{ll}
    (1-a) \frac{n_k(X_{1}^{t})}{t}  & 1 \leq k \leq K(X_{1}^{t}) \\
        a                           & k = K(X_{1}^{t}) + 1 \\
\end{array}.
\right. 
$$

The Simon process strictly follows both Zipf's law and Heaps' law with an exponent of 
1.0. The PY process copes with this problem by decreasing the introduction rate of new 
words in proportion to $K(X_1^t)$ via another parameter $b$, with $ 0 \leq a < 
1$ and $0\leq b$:
$$
\footnotesize
P(x_{t+1} = w_k ) = \left\{
\begin{array}{ll}
     \frac{n_k(X_1^t)-a}{t+b}     &   1 \leq k \leq K(X_1^t) \\
     \frac{a K(X_1^t) + b}{t+b}     & k = K(X_1^t) + 1 \\
  \end{array}.
\right. 
$$
These two parameters would serve to produce Zipf's law with slightly convex behavior 
\citep{goldwater_ml}. The basic models introduced thus far define nothing about how to 
introduce words: we would simply generate random sequences and examine their 
scaling properties, because the basic formulation thus far governs the nature of the 
language model elaborated from these models. By mapping a word to the elements 
produced, however, we would generate a language model, like the two-stage model 
proposed in \citep{goldwater_ml}. Here, we consider a more advanced model 
proposed as the hierarchical Pitman-Yor language model (HPYLM)\citep{teh06}, which 
integrates the Pitman-Yor process into an $n$-gram model.

\begin{figure*}[t]
  \allfig{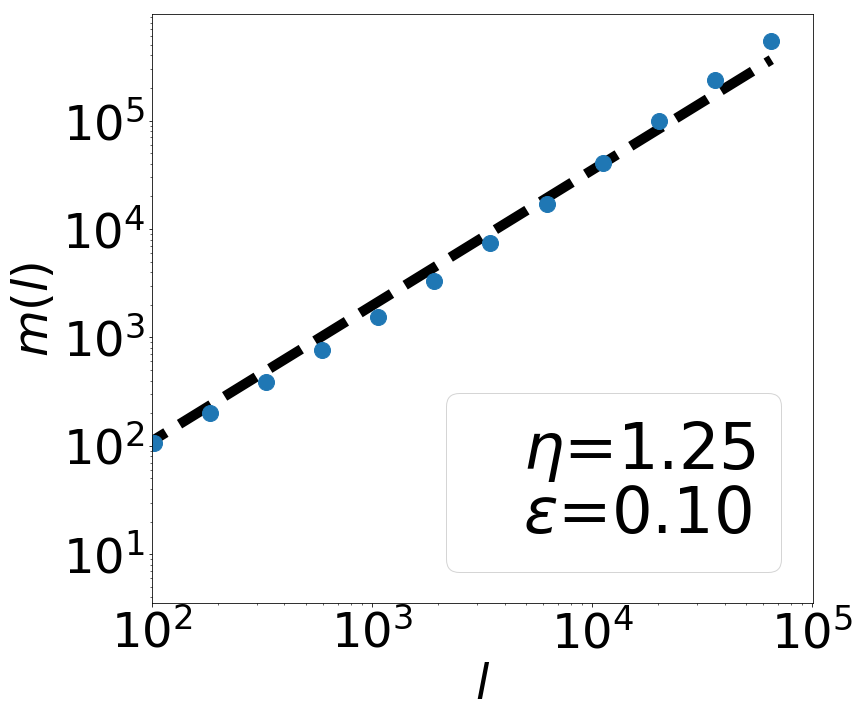}
\vspace*{-0.5cm}
\caption{Scaling properties of AWD-LSTM-Cache model trained with WT2. 
\label{fig:awdcache}}
\end{figure*}

\subsection{Neural Language Models}

The predictive performance of state-of-the-art language models
improved radically with neural language models.
The majority of promising neural language models 
\citep{Mikolov_2012,Melis_2017,Merity_2017,Yang_2017}
adopt recurrent neural networks (RNNs). The RNNs compute a hidden state 
$h_{t}$ from the input $x_{t}$ and the previous hidden state $h_{t-1}$ to incorporate 
past information effectively:
\begin{equation}
h_{t} = \Phi(x_{t},h_{t-1}).
\end{equation}
The function $\Phi$ depends on the recurrent architecture of the network. This article 
focuses on LSTM \citep{Hochreiter_1997}, because the difference in performance is 
insignificant among architectures such as the gated recurrent unit (GRU) 
\citep{Cho_2014} and other LSTM variants 
\citep{Chung_2014,Greff_2015,Melis_2017}. A total of six neural language models 
are  considered.

The first model is an LSTM language model that is not trained with regularization 
techniques. The predictive performance is equivalent or better than with $n$-gram models but 
significantly worse than with state-of-the-art models. This model therefore could be 
considered as a baseline to verify whether further regularization techniques 
contribute to reproducing the scaling properties.

For the advanced models, we adopted averaged stochastic gradient descent (ASGD) 
weight-dropped LSTM, or simply AWD-LSTM \citep{Merity_2017}, because it consists 
of a standard architecture (embedding + LSTM + softmax) yet outperforms many other 
models in terms of perplexity. In addition to the standard AWD-LSTM model, two other 
architectures are considered: continuous cache \citep{Grave_2016} and mixture of 
softmax (MoS) \citep{Yang_2017}.

Continuous cache is a memory augmentation architecture that computes a cache 
probability $p_{cache}$ from context $w_{t-l}^{t}$, where $l$ is a window size 
parameter. It computes the similarity between $h_{t}$ and $h_{i}$ to estimate the 
reappearance of $w_{i}$ at $t+1$. The output probability of the model with continuous 
cache, denoted as the AWD-LSTM-Cache model, is a linear interpolation of the 
AWD-LSTM and the cache probability. We also considered a model 
incorporating the Simon process, denoted as the AWD-LSTM-Simon model. It behaves 
as a uniform sampling from the past generated sequence and is a special case of 
AWD-LSTM-Cache.
MoS reformulates the language model task as matrix factorization and is a 
state-of-the-art language model integrated with AWD-LSTM as the AWD-LSTM-MoS 
model. We also considered a combination of all these architectures, the 
AWD-LSTM-MoS-Cache model.

In our experiments, all of the language models are trained to minimize the negative 
log-likelihood of the training data by stochastic gradient algorithms. The window size 
$l$ for the AWD-LSTM-Simon model is set to 10,000 to balance a large window size 
with computation efficiency.

\begin{table*}[t]
\footnotesize
  \centering
  \caption{Summary of scaling properties of language models with WT2}
  \label{tab:summary-WT2}
  \begin{tabular}{|l||c|c|c|c|c|c|}
    \hline
    &  Perplexity &  \multicolumn{2}{|c|}{Vocabulary Population} & \multicolumn{3}{|c|}{
      Long Memory} \\ \cline{3-7}
   &   & Zipf's & Heaps'  & Ebeling's & Taylor's & Long Range\\
   &             & Law  & Law  　& Method    & Law  & Correlation \\
   &             & $f(r)\propto r^{-\alpha}$ & $v(n) \propto n^\beta$  
   & $m(l) \propto l^{\eta}$ & $\sigma \propto \mu^{\zeta}$ & $c(s) \propto s^{-\xi}$ \\
    \hline
    \multicolumn{7}{|c|}{Original Dataset} \\
    \hline
    Wikitext-2 & - & Yes &  0.75 (0.13) & 1.32 (0.10)  & 0.62 (0.15)  & 0.33 (0.04) \\
    Wikitext-2-raw & - & Yes &  0.78 (0.09) & 1.33 (0.10)  & 0.65 (0.11)  & 0.32 (0.03)\\
    \hline
    \multicolumn{7}{|c|}{$N$-gram Language Model} \\
     \hline
  3-gram  & 195.54 & Yes &  0.78 (0.13) &  1.01 (0.01) & 0.50 (0.02) & No \\
  5-gram with back-off & 181.75 & Yes &   0.78 (0.13)  &  1.00 (0.01) & 0.50 (0.02) & No \\
  \hline
      \multicolumn{7}{|c|}{Grammatical Model} \\
 \hline 
   PCFG(PTB)   &  - &   Yes &  0.73 (0.19) & 1.00 (0.00) & 0.50 (0.02) & No \\
   \hline
 \multicolumn{7}{|c|}{Simon and Pitman-Yor Family} \\
      \hline
Simon ($a=0.1$)  & - & Yes & 0.95 (0.15)   &  - & 0.50 (0.01)  & 0.09 (0.03) \\
 Pitman-Yor ($a=0.8$,$b=1.0$)  & - & Yes & 0.78 (0.09)  &  - & 0.50 (0.01) & No \\
 HPYLM & 187.10 & Yes & 0.78 (0.13) & 1.00 (0.00) & 0.50 (0.02) & No  \\
\hline
    \multicolumn{7}{|c|}{Neural Language Model} \\
 \hline
 LSTM (no regularization)&113.18 & Yes & 0.78 (0.12) & 1.10 (0.03)   & {\bf 0.52} (0.03)  & 0.43 (0.15) \\
 AWD-LSTM           & 64.27 & Yes & 0.76 (0.13) & 1.30 (0.15)  & {\bf 0.58} (0.06)  & 0.05 (0.01) \\ 
 AWD-LSTM-Simon     & 61.59 & Yes & 0.77 (0.10) & 1.25 (0.15)  & {\bf 0.55} (0.05) & 0.03 (0.01) \\
 AWD-LSTM-MoS       & 62.44 & Yes & 0.78 (0.12) & 1.16 (0.07)  & {\bf 0.54} (0.04)  & 0.33 (0.07) \\
 AWD-LSTM-MoS-Cache & 59.21 & Yes & 0.78 (0.11) & 1.20 (0.07)  & {\bf 0.57} (0.07) & 0.29 (0.05) \\
 AWD-LSTM-Cache     & 50.39 & Yes & 0.78 (0.11) & 1.25 (0.10)  & {\bf 0.59} (0.07)  & 0.14 (0.04) \\
 \hline
   \end{tabular}
   \end{table*}

\begin{table*}[t]
\footnotesize
  \centering
  \caption{Summary of scaling properties of language models with PTB}
  \label{tab:summary-PTB}
  \begin{tabular}{|l||c|c|c|c|c|c|}
    \hline
    &  Perplexity &  \multicolumn{2}{|c|}{Vocabulary Population} & \multicolumn{3}{|c|}{
      Long Memory} \\ \cline{3-7}
   &   & Zipf's & Heaps's  & Ebeling's & Taylor's & Long Range\\
   &             & Law  & Law  　& Method    & Law  & Correlation \\
   &             & $f(r)\propto r^{-\alpha}$ & $v(n) \propto n^\beta$  
   & $m(l) \propto l^{\eta}$ & $\sigma \propto \mu^{\zeta}$ & $c(s) \propto s^{-\xi}$ \\
  \hline
    \multicolumn{7}{|c|}{Original Dataset} \\
  \hline
  Penn Tree Bank & - & Yes &  0.70 (0.16) & 1.23 (0.06)  & 0.56 (0.14)  & 0.81 (0.24) \\
  Penn Tree Bank-raw & - & Yes &  0.83 (0.07) & 1.20 (0.05)  & 0.57 (0.06)  & 0.60 (0.16) \\
  \hline
  \multicolumn{7}{|c|}{$N$-gram Language Model} \\
  \hline    
  3-gram  & 126.71 & Yes &   0.71 (0.19) &  1.00 (0.00) & 0.50 (0.02) & No \\
  5-gram with back-off & 112.87 & Yes & 0.71 (0.19) &  1.01 (0.01)  & 0.50 (0.02) & No \\
  \hline
      \multicolumn{7}{|c|}{Grammatical Model} \\
 \hline
   PCFG   &  - &   Yes &  0.73 (0.19) & 1.00 (0.00) & 0.50 (0.02) & No \\
   \hline
 \multicolumn{7}{|c|}{Simon and Pitman-Yor Family} \\
      \hline
 HPYLM& 144.41 & Yes & 0.73 (0.21) & 1.00 (0.00) & 0.50(0.02) & No  \\
\hline
    \multicolumn{7}{|c|}{Neural Language Model} \\
 \hline
LSTM (no regularization)&111.79 & Yes & 0.71 (0.19) & 1.04 (0.01) & {\bf 0.51} (0.02)  & 0.84 (Weak) \\
AWD-LSTM           & 56.40 & Yes & 0.71 (0.18) & 1.06 (0.02)  & {\bf 0.51} (0.03)  & 0.69 (Weak) \\ 
AWD-LSTM-Simon     & 57.85 & Yes & 0.72 (0.16) & 1.04 (0.01)  & {\bf 0.51} (0.03)  & No \\
AWD-LSTM-MoS       & 54.77 & Yes & 0.71 (0.18) & 1.10 (0.03)  & {\bf 0.52} (0.04)  & 0.77 (Weak) \\
AWD-LSTM-MoS-Cache & 54.03 & Yes & 0.71 (0.18) & 1.13 (0.04) & {\bf 0.55} (0.06) & 0.61 (Weak) \\
AWD-LSTM-Cache     & 52.51 & Yes & 0.72 (0.17) & 1.07 (0.02) & {\bf 0.53} (0.05)  & 0.57 (Weak) \\
\hline
  \end{tabular}
\end{table*}

\section{Scaling Properties of Language Models}

We chose two standard datasets, WT2 and PTB, for training language
models. \tabref{tab:summary-WT2} and \tabref{tab:summary-PTB} list the
overall results for WT2 and PTB, respectively. Each table contains
five blocks, with the first indicating the properties of the original
datasets with and without preprocessing. The remaining blocks indicate
the results for the language models. Every row gives the results for a
generated text of 1 million words. The two rows for the Simon and PY
processes appear only in the table for WT2, because they represent
non-linguistic random sequences obtained by the definitions given in
\secref{sec:py}, and no real data learning is involved. All the other
rows show the results of the corresponding language model learning
either the WT2 or PTB dataset. The datasets were preprocessed to
reduce the vocabulary size. Infrequent words were replaced with
\verb!<unk>!, and numbers were replaced with \verb!N! in PTB.


The first column lists the perplexity measured by the model. For some
models, perplexity does not apply, as indicated by the symbol
``-''. The perplexity grows with the Simon and PY models, and that of
the PCFG cannot be compared under the same criteria.

For the Zipf's law column, when the law holds, this is indicated by
``Yes''. The reason is that, as mentioned in \secref{sec:zipf}, the
exponent is often difficult to estimate. Ebeling's method is not
applicable to the Simon and PY processes without a word generation
model. For the other models, when an exponent was obtained, we checked
whether the value was larger than 1.0 (the i.i.d. case). Similarly,
for Taylor's law, we checked whether the exponent was larger than
0.5. If the text is not long-range correlated, this is mentioned by
``No'' or ``Weak'': ``No'' if more than one value was negative for
$s\leq10$, or ``Weak'' for $s\leq100$.

The two tables do not indicate qualitative differences except for the
cells in the last block of the last column: for other cells, when a
scaling property held for one, it also held for the other. The main
difference occurred because of a difference in dataset quality: PTB
(The Wall Street Journal) consists of short articles, whereas WT2
(Wikipedia) consists of longer articles written coherently with
careful preprocessing. WT2 therefore has stronger long memory than
PTB, which is apparent from their long memory properties. 

\begin{figure}[t]
\centering
\includegraphics[width=\columnwidth]{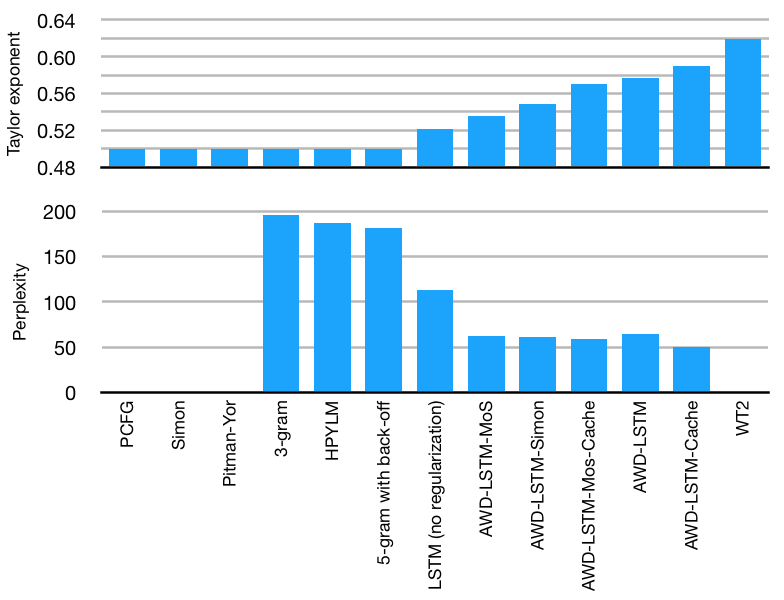}
\vspace*{-1cm}
\caption{
Values of the Taylor exponent and perplexity for all the
  language models and the real data of WT2. When the perplexity is
  unmeasurable (for the original WT2 data and some language
  models), no bar appears. The results show that the two measures are
  correlated, but not totally among the most advanced models.}
\label{fig:bars}
\vspace*{-0.5cm}
\end{figure}

The language models listed above the neural language models in the
tables failed to reproduce all of the scaling properties for long
memory. \figref{fig:hpylm} shows a set of graphs for the HPLYM: apart
from Zipf's and Heaps' laws, this model did not exhibit any long
memory. The $n$-gram models and PCFG had similar tendencies. The sole
exception was the Simon model, which presented strong long-range
correlation, even stronger than that of the original WT2.
Its Taylor exponent, however, was 0.5, suggesting its essential
difference from a natural language text. This is obvious, in a way,
because a sequence produced by the Simon model has a different
mathematical nature with respect to vocabulary population for small
$t$ and large $t$, which is not the case for natural
language. Curiously, this long memory was destroyed with the PY models
and HPLYM. Overall, the Simon- and PY-based modeling might not be
adequate for natural language.

In contrast, for WT2, the neural language models were able to
reproduce all of the scaling properties. \figref{fig:awdcache} shows a
set of graphs for the AWD-LSTM-Cache model, for WT2. The figure shows
how well the model captures the nature of the original text in terms
of every property.

The analysis, however, suggests the possibility of further
improvement. Even though the AWD-LSTM-Cache model performed the best
for WT2 with respect to perplexity, its Taylor exponent $\zeta$ still
remained smaller (0.59) than that of the original text
(0.62). Verification of the actual generated text showed that this
model had a tendency to repeat words locally. Moreover, the neural
language models failed to reproduce long memory for the PTB
dataset. Although the long memory underlying the original PTB is
weaker than that of WT2, PTB still has long memory, and yet the models
could only reproduce ``weak'' long-range correlation. This and all the
above findings suggest directions for improving the neural language
models.

At the same time, the discussion highlights the importance of the
dataset used for evaluation: PTB is less suitable for evaluating long
memory, while WT2 is recommended for this purpose.

\section{Effectiveness of Scaling Properties for Examining Language Models}

The results thus far also enable discussion from another viewpoint, of
considering which scaling property would most effectively evaluate
language models. Most important is the question in the article's
title: ``Is perplexity sufficient?''. We might ask whether model
quality correlates with proximity to the exponent for a dataset, and
which scaling property is the most effective for evaluation. We thus
examine every property from this viewpoint.

First, Zipf's law does not add exploitable information on the quality
of a language model, because its exponent is difficult to estimate, as
discussed in \secref{sec:zipf}, and because all the models exhibited
behavior like Zipf's law. Still, the Zipf's law distribution for WT2
in \figref{fig:phy} showed a drop at the tail, indicating that
something unnatural occurred (i.e., replacement of rare words with
\verb!<unk>!). Therefore, Zipf's law could be applied just to verify
whether a vocabulary population is normal. Since Heaps' law is derived
from Zipf's law, it also does not contribute much to evaluating
language models.

Turning to long memory, the three properties roughly have positive
correlation: when the original dataset had weaker long memory (PTB as
compared with WT2), the degree of memory exhibited by the neural language
models was also weaker. Ebeling's method is limited in that it only
applies for characters. The long-range correlation also has a limited
capacity, because the Simon model exhibited strong long-range
correlation but had a Taylor exponent of 0.5. Taylor's law therefore
seems more credible for evaluating model quality. \figref{fig:bars}
shows the values of the Taylor exponent and perplexity for every
language model. The figure indicates that they correlated well with
differences for the most advanced neural language models. It
highlights some aspects of the model characteristics. For example, it
is likely that the long memory quality for WT2 was degraded by MoS,
even though it improved the perplexity from AWD-LSTM. These results
demonstrate that evaluation of language models from multiple
viewpoints would contribute to better understanding of the nature of
learning techniques, and scaling properties can provide these
different viewpoints.

\section{Conclusion}

We explored the performance of language models with respect to the
scaling properties of natural language and proposed evaluation methods
other than perplexity. We listed five such tests, two for vocabulary
population and three for long memory. For vocabulary population, all
models considered here presented the scaling property. In contrast,
many models did not exhibit long memory, but the state-of-the-art
neural language models were able to produce a sequence with long
memory. This constitutes solid evidence of the difference between
$n$-gram and neural language models. Moreover, the difference between
the exponents from a model and from the original dataset shows the
possibility of improvement.

Further verification along the same line using raw data and other
datasets remains for our future work. We also intend to investigate
other kinds of metrics for model assessment.

\bibliographystyle{natbib}
\bibliography{main}

\end{document}